\documentclass{article}
\usepackage{RCRC_arxiv_eng25}
\usepackage[utf8]{inputenc} 
\usepackage[T1]{fontenc}    
\usepackage{booktabs}       
\usepackage{amsfonts}       
\usepackage{nicefrac}       
\usepackage{microtype}      

\usepackage[pdftex]{graphicx}
\usepackage{amsmath,amssymb}
\usepackage{mathtools}
\mathtoolsset{showonlyrefs=true}
\usepackage{cite}
\title{Convolutional Reservoir Computing for World Models}

\author{
  Hanten~Chang \\
  Graduate school of Systems and \\Information Engineering,\\
  University of Tsukuba, Japan\\
  \texttt{s1820554@s.tsukuba.ac.jp} \\
   \And
 Katsuya~Futagami \\
Graduate school of Systems and \\Information Engineering,\\
  University of Tsukuba, Japan\\
  \texttt{s1820559@s.tsukuba.ac.jp} \\
}

\begin{document}
\maketitle
\begin{abstract}
Recently, reinforcement learning models have achieved great success, completing complex tasks such as mastering Go and other games with higher scores than human players. Many of these models collect considerable data on the tasks and improve accuracy by extracting visual and time-series features using convolutional neural networks (CNNs) and recurrent neural networks, respectively. However, these networks have very high computational costs because they need to be trained by repeatedly using a large volume of past playing data. In this study, we propose a novel practical approach called reinforcement  learning with convolutional reservoir computing (RCRC) model. The RCRC model has several desirable features: 1. it can extract visual and time-series features very fast because it uses random fixed-weight CNN and the reservoir computing model; 2. it does not require the training data to be stored because it extracts features without training and decides action with evolution strategy. Furthermore, the model achieves state of the art score in the popular reinforcement learning task. Incredibly, we find the random weight-fixed simple networks like only one dense layer network can also reach high score in the RL task.
\end{abstract}


\section{Introduction}
Recently, reinforcement learning (RL) models have achieved great success, mastering complex tasks such as Go\cite{silver2016mastering, silver2017mastering} and other games\cite{DBLP:journals/corr/MnihKSGAWR13, DBLP:journals/corr/abs-1803-00933, kapturowski2018recurrent} with higher scores than human players. Many of these models use convolutional neural networks (CNNs) to extract visual features directly from the environment state pixels\cite{DBLP:journals/corr/abs-1708-05866}. Some models use recurrent neural networks (RNNs) to extract time-series features and achieved higher scores\cite{kapturowski2018recurrent, hausknecht2015deep}. 

However, these deep neural network (DNN) based models are very computationally expensive in that they train networks weights by repeatedly using a large volume of past playing data. Certain techniques can alleviate these costs, such as the distributed approach\cite{DBLP:journals/corr/abs-1803-00933, mnih2016asynchronous} which efficiently uses multiple agents, and the prioritized experienced replay\cite{schaul2015prioritized} which selects samples that facilitate training. However, the cost of a series of computations, from data collection to action determination, remains high.

World model\cite{DBLP:journals/corr/abs-1803-10122, NIPS2018_7512} can also reduce computational costs by completely separating the training process between the feature extraction model and the action decision model.  World model replaces the feature extraction model training process with the supervised learning, by using variational auto-encoder (VAE)\cite{2013arXiv1312.6114K, 2014arXiv1401.4082J} and mixture density network combined with an RNN (MDN-RNN) \cite{DBLP:journals/corr/Graves13, ha2017recurrent}. After extracting the environment state features, it uses an evolution strategy called the covariance matrix adaptation evolution strategy (CMA-ES)\cite{doi:10.1162/106365601750190398, DBLP:journals/corr/Hansen16a} to train an action decision model, which achieved outstanding scores in popular RL tasks. The separation of these two models results in the stabilization of feature extraction and omission of parameters to be trained based on task-dependent rewards. 

From the success of world model, it is implied that in the RL feature extraction process, it is necessary to extract the features that express the environment state rather than features trained to solve the tasks. In this study, adopting this idea, we propose a new method called "reinforcement learning with convolutional reservoir computing (RCRC)". The RCRC model is inspired by the reservoir computing. 

Reservoir computing\cite{verstraeten2007experimental, lukovsevivcius2009reservoir} is a kind of RNNs, but the model weights are set to random. One of the models of the reservoir computing model, the echo state network (ESN)\cite{jaeger2001echo, jaeger2004harnessing, lukovsevivcius2012practical} is used to solve time-series tasks such as future value prediction. For this, the ESN extracts features for the input based on the dot product between the input and a random matrix generated without training. Surprisingly, features obtained in this manner are expressive enough to understand the input signal, and complex tasks such as chaotic time-series prediction can be solved by using them as the input for a linear model. In addition, the ESN has solved the tasks in multiple fields such as time-series classification \cite{tanisaro2016time, ma2016functional} and Q-learning-based RL\cite{szita2006reinforcement}. Thus, even if the ESN uses random weights, it can extract sufficient expressive features of the input and can solve the task using the linear model. Similarly, in image classification, the model that uses features extracted by the CNN with random fixed-weights as the ESN input achieves high accuracy classification with a smaller number of parameters\cite{Tong2018}.

Based on the success of the above random fixed-weight models, RCRC extracts the visual features of the environment state using random fixed-weight CNN and, using these features as the ESN input, extracts time-series features of the environment state transitions. In the feature extraction process, all features are extracted based on matrices with random elements. Therefore, no training process is required, and feature extraction can be performed very fast. After extracting the environment state features, we use CMA-ES\cite{DBLP:journals/corr/Hansen16a, doi:10.1162/106365601750190398} to train a linear combination of extracted features to perform the actions, as in world model. This model architecture results in the omission of the training process of feature extraction and recuded computational costs; there is also no need to store a large volume of past playing data. Furthermore, we show that RCRC can achieve state of the art score in popular RL task. 

Our contribution in this paper is as follow:
\begin{itemize}
\item We developed a novel and highly efficient approach to extract visual and time-series features of an RL environment state using a fixed random-weight model with no training. 
	\item By combining random weight networks with an evolution strategy method, we eliminated the need to store any past playing data. 
	\item We showed that a model with these desirable characteristics can achieve state of the art score in popular continuous RL task. 
	\item We showed that simple random weight-fixed networks, for example one dense layer network, can also extract visual features and achieve high score in continuous RL task. 
\end{itemize}

\section{Related Work}
\subsection{Reservoir Computing}
\begin{figure}[t]
  \centering
  \includegraphics[keepaspectratio, width=12cm, clip]{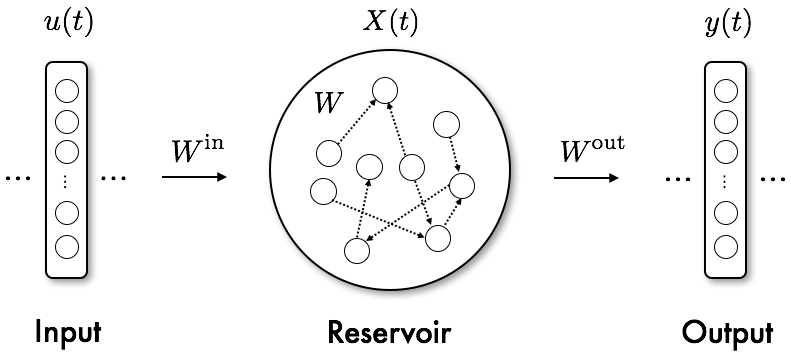}
  \caption{Reservoir Computing overview for the time-series prediction task.}
  \label{fig:fig1}  
\end{figure}
Reservoir computing is a promising model that can solve complex tasks, such as chaotic time-series prediction, without training for the feature extraction process. In this study, we focus on the reservoir computing model, ESN\cite{jaeger2001echo, jaeger2004harnessing, lukovsevivcius2012practical}. ESN was initially proposed to solve time-series tasks\cite{jaeger2001echo} and is regarded as an RNN model\cite{lukovsevivcius2009reservoir}, it can be applied to multiple fields. 

Let the $N$-length, $D_u$-dimensional input signal be $u = \{u(1), u(2), ..., u(t),...,u(N)\} \in \mathbb{R}^{N \times D_u}$ and the signal that adds input signal to one bias term be $U = [u;1] = \{U(1),U(2),...,U(T),...,U(N)\} \in \mathbb{R}^{N \times (D_u+1)}$. [;] is a vector concatenation. ESN gets features called the reservoir state $X = \{X(1),, ..., X(t),...,X(N)\} \in \mathbb{R}^{N \times D_x}$ as follows: \begin{eqnarray}
	\tilde{X}(t+1) & = & f(W^{\text{in}}U(t) + WX(t)))\\
	X(t+1) & = & (1-\alpha)X(t) + \alpha\tilde{X}(t+1) 
\end{eqnarray}
where the matrices $W^{\text{in}} \in \mathbb{R}^{(D_u+1) \times D_x}$ and $W \in \mathbb{R}^{D_x \times D_x}$ are random sampled from a probability distribution such as a Gaussian distribution, and $f$ is the activation function which is applied element-wise. As the activation function, $linear$ and $tanh$ functions are generally used; it is also known that changing the activation function according to the task improves accuracy\cite{inubushi2017reservoir, DBLP:journals/corr/abs-1905-09419}. The leakage rate $\alpha \in [0,1]$ is a hyper parameter that tunes the weight between the current and the previous values, and $W$ has two major hyper parameters called sparsity which is ratio of 0 elements in matrix $W$ and the spectral radius that is memory capacity hyper parameter which is calculated by the maximal absolute eigenvalue of $W$. 

Finally, ESN estimates the target signal $y = \{y(1), y(2), ..., y(t),...,y(N)\} \in \mathbb{R}^{N \times D_y}$ as \begin{eqnarray}
y(t) = W^{\text{out}} [X(t); U(t); 1].\end{eqnarray}
 The weight matrix $W^{\text{out}} \in \mathbb{R}^{D_y \times (D_x + D_u + 1)}$ is estimated by a linear model such as ridge regression. An overview of reservoir computing is shown in Figure\ref{fig:fig1}.
 
The unique feature of the ESN is that the two matrices $W ^{\text{in}}$ and $W$ used to update the reservoir state are randomly generated from a probability distribution and fixed without training. Therefore, the training process in the ESN consists only of a linear model to estimate $W ^{\text{out}}$; therefore, the ESN model has a very low computational cost. In addition, the reservoir state reflects complex dynamics despite being obtained by random matrix transformation, and it is possible to use it to predict complex time-series by simple linear transformation\cite{verstraeten2007experimental, jaeger2001echo, DBLP:journals/corr/GoudarziBLTS14}. Because of the low computational cost and high expressiveness of the extracted features, the ESN is also used to solve  other tasks such as time-series classification\cite{tanisaro2016time, ma2016functional}, Q-learning-based RL\cite{szita2006reinforcement} and image classification\cite{Tong2018}.
 
\subsection{World models}
Recently, most RL models use DNNs to extract features and solved several complex tasks. However, these models have high computational costs because a large volume of past playing data need to be stored, and network parameters need to be updated using the back propagation method. There are certain techniques\cite{DBLP:journals/corr/abs-1803-00933, mnih2016asynchronous,schaul2015prioritized} and models that can reduce this cost; some models\cite{DBLP:journals/corr/abs-1803-10122, NIPS2018_7512, DBLP:journals/corr/abs-1811-04551} separate the training process of the feature extraction and action decision models to more efficiently train the action decision model. 

The world model\cite{DBLP:journals/corr/abs-1803-10122, NIPS2018_7512} is one such model, and uses VAE\cite{2013arXiv1312.6114K, 2014arXiv1401.4082J} and MDN-RNN\cite{DBLP:journals/corr/Graves13, ha2017recurrent} as feature extractors. They are trained using supervised learning with randomly played 10000 episodes data. As a result, in the feature extraction process, the task-dependent parameters are omitted, and there remains only one weight parameter to be trained that decides the action in the model. Therefore, it becomes possible to use the evolution strategy algorithm CMA-ES\cite{doi:10.1162/106365601750190398, DBLP:journals/corr/Hansen16a} efficiently to train that weight parameter. The process of optimizing weights of action decision model using CMA-ES can be parallelized. Although the feature extraction model is trained in a task-independent manner, world model achieved outstanding scores and masterd popular RL task {\tt CarRacing-v0}\cite{carracingv0}.

CMA-ES is one of the evolution strategy methods used to optimize parameters using a multi-candidate search generated from a multivariate normal distribution $\mathcal{N}(m, \sigma^2C)$. The parameters $m$, $\sigma$, and $C$ are updated with a formula called the evolution path. Evolution paths are updated according to the previous evolution paths and evaluation scores. Because CMA-ES updates parameters using only the evaluation scores calculated by actual playing, it can be used regardless of whether the actions of the environment are continuous or discrete values\cite{doi:10.1162/106365601750190398, DBLP:journals/corr/Hansen16a}. Furthermore, training can be faster because the calculations can be parallelized by the number of solution candidates. In world model, the action decision model is simplified to reduce the number of task-dependent parameters, making it possible to use CMA-ES efficiently\cite{DBLP:journals/corr/abs-1803-10122, NIPS2018_7512}.

World model improves the computational cost of the action decision model and accelerates the training process by separating models and applying CMA-ES. However, in world model, it is necessary to independently optimize VAE, MDN-RNN, and CMA-ES. Further, because the feature extraction model is dependent on the environment, a large amount of past data must be stored to train the feature extraction model each time the environment changes.

\begin{figure}[t]
  
  \includegraphics[keepaspectratio, width=16.5cm, clip]{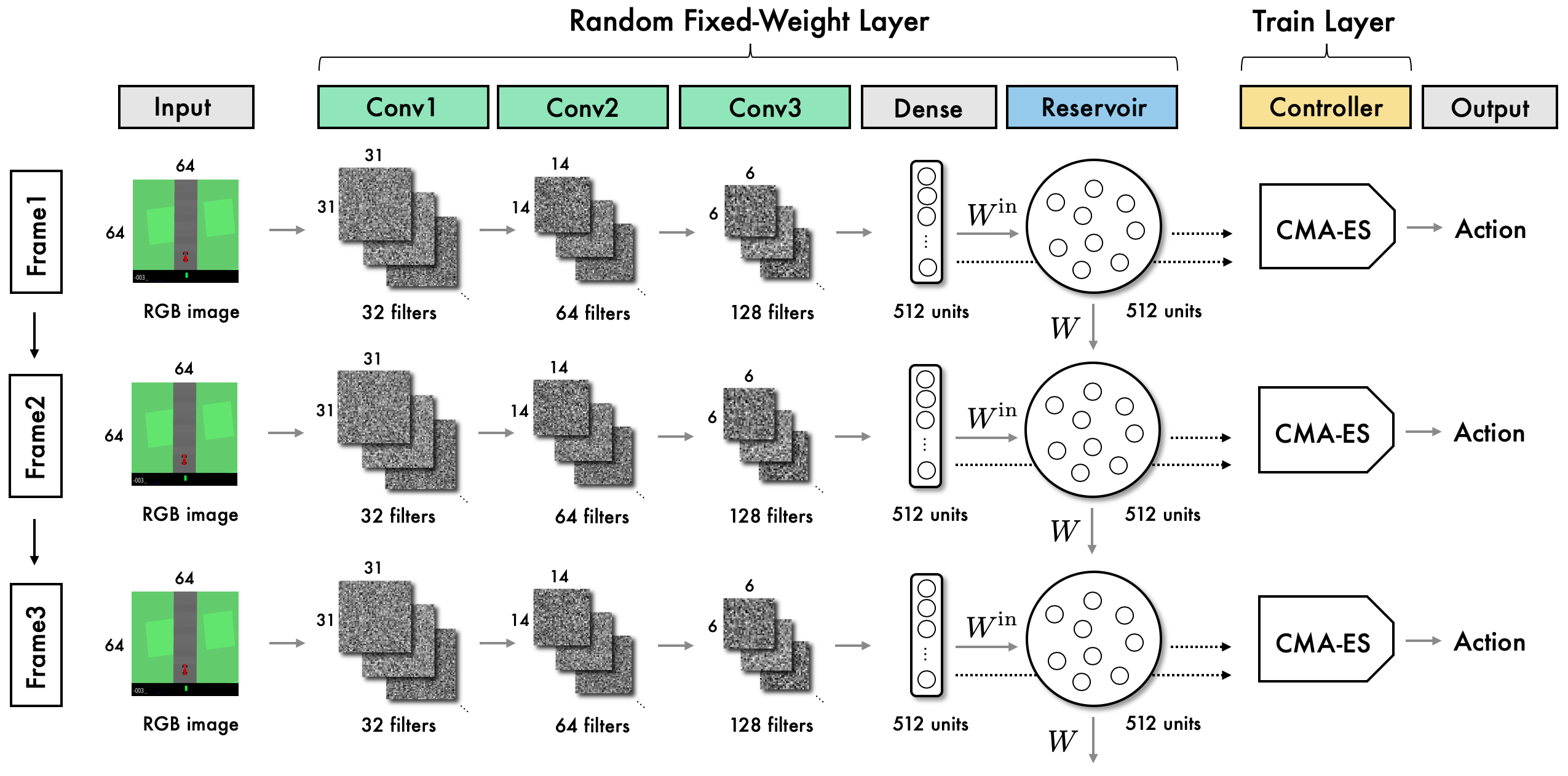}
  \centering
  \caption{RCRC overview to choose the action for {\tt CarRacing-v0}: the first and second layers are collectively called the convolutional reservoir computing layer, and both layers' model weights are sampled from Gaussian distribution and then fixed. }
  \label{fig:fig2} 
\end{figure}
\section{Proposal Model}
\subsection{Basic Concept}
World model\cite{DBLP:journals/corr/abs-1803-10122, NIPS2018_7512} extracts visual features and time-series features of environment states by using VAE\cite{2013arXiv1312.6114K, 2014arXiv1401.4082J} and MDN-RNN\cite{DBLP:journals/corr/Graves13, ha2017recurrent} without using environment scores. The models achieve outstanding scores through the linear transformation of these features. This implies that it only requires features that sufficiently express the environment state, rather than features trained to solve the task.

We thus focus on extracting features that sufficiently express environment state by networks with random fixed-weights. Using networks whose weights are random and fixed has some advantages, such as having very low computational costs and no data storage requirements, while being able to sufficiently extract features. For example, a simple CNN with random fixed-weights can extract visual features and achieve high accuracy\cite{Tong2018}. Although the MDN-RNN is fixed in the world model, it can achieve outstanding scores\cite{Corentin2018rnnfix}. In the case of ESN, the model can predict complex time-series using features extracted by using random matrices transformations\cite{verstraeten2007experimental, jaeger2001echo,DBLP:journals/corr/GoudarziBLTS14}. Therefore, it can be considered that CNN can extract visual features and ESN can extract time-series features, even if their weights are random and fixed. From this hypothesis, we propose reinforcement learning with the RCRC model, which includes both random fixed-weight CNN and ESN.
 
\subsection{Proposal model overview}
The RCRC model is divided into three model layers. In the first layer, it extracts visual features by using a random fixed-weight CNN. In the second layer, it uses a series of visual features extracted in the first layer as input to the ESN to extract the time-series features. In the two layers above, collectively called the convolutional reservoir layer, visual and time-series features are extracted with no training. In the final layer, the linear combination matrix is trained from the outputs of the convolutional reservoir layer to the actions. An overview is shown in Figure\ref{fig:fig2}.

In the previous study, there is a similar world model--based approach\cite{DBLP:journals/corr/abs-1906-08857} that uses fixed weights in VAE and a memory component based on recurrent long short-term memory (LSTM)\cite{hochreiter1997long}. However, this approach is ineffective in solving {\tt CarRacing-v0}. In the training process, the best average score over 20 randomly created tracks of each generation were less than 200. However, as mentioned further on, we achieve an average score above 900 over 100 randomly created tracks by taking reservoir computing knowledge in the RCRC model.

The characteristics of the RCRC model are as follows:
\begin{itemize}
\item The computational cost of this model is very low because visual and time-series features of game states are extracted using a convolutional reservoir computing layer whose weights are fixed and random.
\item In RCRC, only a linear combination in the controller layer needs to be trained because the feature extraction model (convolutional reservoir computing layer) and the action training model (controller layer) are separated.
\item RCRC can take a wide range of actions regardless of continuous or discrete,  because of maximizing the scores that is measured by actually playing.
\item Past data storage is not required, as neither the convolutional reservoir computing layer nor the controller layer need to repeatedly train the past data as in backpropagation.
\item The convolutional reservoir computing layer can be applied to other tasks without further training, because the layer is fixed with task-independent random weights.
\end{itemize}

\subsection{Convolutional Reservoir Computing layer}

In the convolutional reservoir computing layer, the visual and time-series features of the environment state image are extracted by a random fixed-weight CNN and an ESN which has random fixed-weight, respectively. A study using CNN with fixed random weights for each single-image as input to ESN has been previously conducted, and has shown its ability to classify MNIST dataset\cite{mnistdata} with high accuracy\cite{Tong2018}. Based on this study, we developed a novel approach to perform RL tasks. By taking advantage of the RL characteristic by which the current environment state and action determine the next state, RCRC updates the reservoir state with current and previous features. This updating process enables the reservoir state to have time-series features.

More precisely, consider the $D_\text{conv}$-dimensional visual features extracted by fixed random weight CNN for $t$-th environment state pixels $X_{\text{conv}}(t) \in \mathbb{R}^{D_\text{conv}}$ and the $D_\text{esn}$-dimensional reservoir state $X_{\text{esn}}(t) \in \mathbb{R}^{D_\text{esn}}$. The reservoir state $X_{\text{esn}}$ is time-series features and is updated as follows:
\begin{eqnarray}
\tilde{X}_{\text{esn}}(t+1) & = & f(W^{\text{in}}X_{\text{conv}}(t) + WX_{\text{esn}}(t)))\\
X_{\text{esn}}(t+1) & =&  (1-\alpha)X_{\text{esn}}(t) + \alpha\tilde{X}_{\text{esn}}(t+1).
\end{eqnarray}
This updating process has no training requirement, and is very fast, because $W^{\text{in}}$ and $W$ are random matrices sampled from the probability distribution and fixed. 

\subsection{Controller layer}
The controller layer decides the action by using the output of the convolutional reservoir computing layer, $X_{\text{conv}}$ and $X_{\text{esn}}$. Let $t$-th environment state input vector which added one bias term be $S(t) = [X_{\text{conv}}(t); X_{\text{esn}}(t); 1]\in \mathbb{R}^{D_{\text{conv}}+D_{\text{esn}}+1})$. In the action decision, we suppose that the feature $S(t)$ has sufficient expressive information and it can take action by a linear combination of $S(t)$. Therefore, we obtain action $A(t) \in \mathbb{R}^{N_{\text{act}}}$ as follows:
\begin{eqnarray}
  \tilde{A}(t) & = & W^\text{out}S(t)\\
  A(t) & = & g(\tilde{A}(t)) 
\end{eqnarray}
where, $W^{\text{out}} \in \mathbb{R}^{(D_{\text{conv}}+D_{\text{esn}}+1) \times N_{\text{act}}}$ is the weight matrix and $N_{\text{act}}$ is the number of actions in the task environment; $g$ is applied to each action to put each $\tilde{A}(t)$ in the range of possible values in the task environment.

Because the weights of the convolutional reservoir computing layer are fixed, only the weight parameter $W^{\text{out}}$ requires training. We optimize $W^{\text{out}}$ by using CMA-ES, as in world model. Therefore, it is possible to parallelize the training process and handle both discrete and continuous values as actions\cite{doi:10.1162/106365601750190398, DBLP:journals/corr/Hansen16a}. 

The process of optimizing $W^\text{out}$ by CMA-ES are shown as follows:
\begin{enumerate}
  \item  Generate $n$ solution candidates $W_{T,i}^{\text{out}} (i=1,2,...,n)$ from a multivariate normal distribution $\mathcal{N}(m(T), \sigma(T)^2C(T))$
  \item  Create $n$ environments and agents $\text{worker}_i$ that implement RCRC 
  \item  Set $W_i^{\text{out}}$ to the controller layer of $\text{worker}_i$ 
  \item  In each execution environment, each $\text{worker}_i$ plays $m$ episodes and receives $m$ scores $G_{i, j}(j =1, 2, ..., m)$
  \item  Update evolution paths with the score of each $W_i^{\text{out}}$ which is $G_i = 1/m\sum_{j=1}^m{G_{i, j}}$ 
  \item  Update $m$, $\sigma$, $C$ using evolution paths
  \item  Generate a new $n$ solution candidate $W_{(T+1),i}^{\text{out}}(i=1, 2, ..., n)$ from the updated multivariate normal distribution $\mathcal{N}(m(T+1), \sigma(T+1)^2C(T+1))$
  \item  Repeat 2 to 7 until the convergence condition is satisfied or the specified number of repetitions are completed
\end{enumerate}
In this process, $T$ represents an update step of the weight matrix $W^{\text{out}}$, and $n$ is the number of solution candidates $W^{\text{out}}$ generated at each step. The worker is an agent that implements RCRC, and each worker extracts features, takes the action and plays in each independent environment to obtain scores.  Therefore, it is possible to parallelize $n$ processes to calculate each score.

\section{Experiments}
\subsection{CarRacing-v0}
\begin{figure}[t]
  \centering
  \includegraphics[keepaspectratio, width=8cm, clip]{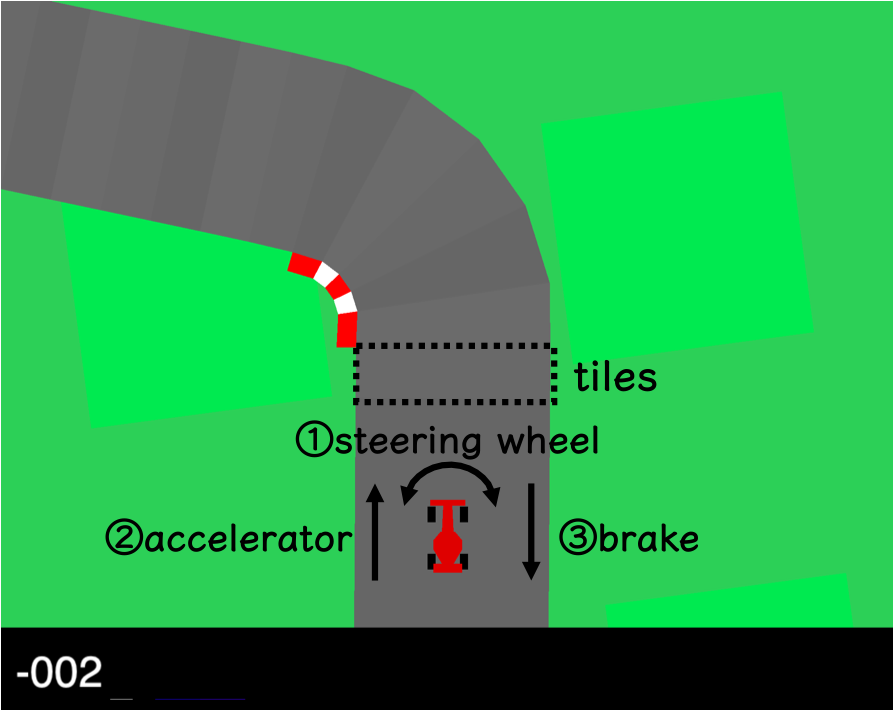}
  \caption{Example environment state image of {\tt CarRacing-v0} and three parameters in the enviroments. The score is added when the car passes through a tile laid on the course.}
  \label{fig:fig3} 
\end{figure}

We evaluate the RCRC model in the popular RL task {\tt CarRacing-v0}\cite{carracingv0} in OpenAI Gym\cite{1606.01540}. This is a car racing game environment that was known as a difficult continuous actions task\cite{DBLP:journals/corr/abs-1803-10122, NIPS2018_7512}. The goal of this game is to go around the course without getting out by operating a car with three continuous parameters: steering wheel $[-1, 1]$, accelerator $[0, 1]$, and brake $[0, 1]$. Each frame of the game screen is given by RGB 3 channels and 96 $\times$ 96 pixels. The course is filled with tiles as shown in Figure\ref{fig:fig3}. Each time the car passes a tile on the course, $1000 / N$ is added to the score. $N$ is the total number of tiles on the course. The course is randomly generated each time, and the total number of tiles in the course varies around 300. If all the tiles are passed, the total reward will be 1000, but it is subtracted by 0.1 for each frame. The episode ends when all the tiles are passed or when 1000 frames are played. If the player can pass all the tiles without getting out of the course, the reward will be over 900. The definition of "solve" in this game is to get an average of 900 per 100 consecutive trials.

\subsection{Precedure}
In the convolutional reservoir computing layer, we set 3 convolution layers and 1 dense layer. The filter sizes in the convolution layers are 31, 14, and 6, and the strides are all 2. We set $D_\text{conv}$ and $D_\text{esn}$ to 512 to expand the features. In the reservoir computing layer, we also set the sparsity of $W$ to 0.8; the spectral radius of $W$ to 0.95. All activation functions are set to $tanh$, which is often used in reservoir computing and achieves higher scores.

As in world model, we set three units ($\tilde{A}_{\text{1}}(t)$, $\tilde{A}_{\text{2}}(t)$, and $\tilde{A}_{\text{3}}(t)$) as output of the controller layer, and each of them corresponds to an action: steering wheel $A_{\text{1}}(t)$, accelerator $A_{\text{2}}(t)$, and brake $A_{\text{3}}(t)$ \cite{DBLP:journals/corr/abs-1803-10122, NIPS2018_7512}. Also as in world model, each action $A(t)$ is determined by converting each $\tilde{A}(t)$ by $g$ shown as follows\cite{DBLP:journals/corr/abs-1803-10122, NIPS2018_7512}: 
\begin{eqnarray}
g(\tilde{A}(t)) = 
\begin{cases}
    tanh\left(\tilde{A}_{\text{1}}(t)\right) \\
    [tanh\left(\tilde{A}_{\text{2}}(t)\right) + 1.0] / 2.0 \\
    clip[tanh\left(\tilde{A}_{\text{3}}(t)\right), 0, 1]
\end{cases}
.
\end{eqnarray}
The function $clip[x, \lambda_{\text{min}}, \lambda_{\text{max}}]$ is a function that limits the value of $x$ in range from $\lambda_{\text{min}}$ to $\lambda_{\text{max}}$ by clipping. 

In the experiment, 16 workers $(n=16)$ with different $W^{\text{out}}$ parameters are prepared for each update step, and each worker is set to simulate over 8 randomly generated tracks $(m=8)$, and update $W^{\text{out}}$ with an average of these scores.  As the input value, each frame is resized to 3 channels of 64 $\times$ 64 pixels. As in world model\cite{DBLP:journals/corr/abs-1803-10122, NIPS2018_7512}, we evaluate an average score over 100 randomly created tracks score as the generalization ability of the models. 

To investigate the ability of each network structures, we evaluate three models: full RCRC model, the RCRC model that removes the reservoir computing layer from convolutional reservoir computing layer (visual model), the RCRC model that has only one dense layer as feature extractor (dense model). The dense model uses flatten vector of 64$\times$64 with 3 channels image, and weights of all models are random and fixed. The visual model extracts visual features and the dense model extracts only visual features with no convolutional process. In both the visual model and the dense model, the inputs to the controller layer are the $D_{\text{conv}}$-dimensional outputs from the dense layer shown in Figure\ref{fig:fig2}, and one bias term.

\begin{figure}[t]
  \centering
  \includegraphics[keepaspectratio, width=14cm, clip]{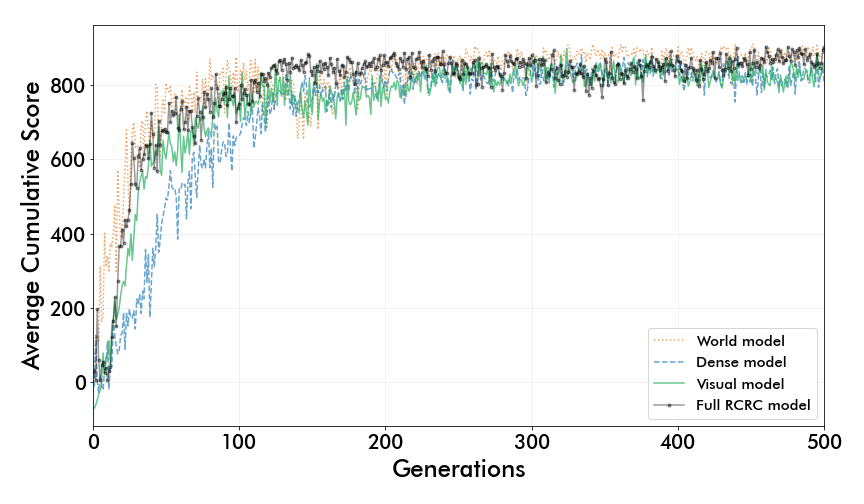}
  \caption{The best average score over 8 randomly created tracks among 16 workers at {\tt CarRacing-v0}.}
  \label{fig:fig4} 
\end{figure}
\begin{table}
 \caption{{\tt CarRacing-v0} scores of various methods.}
  \centering
  \begin{tabular}{lll}
    \toprule
    Method     & Average Score  \\
    \midrule
    DQN\cite{DQNscore} & 343  $\pm$ 18     \\  
    DQN + Dropout\cite{dqndropscore} & 892  $\pm$ 41     \\      
    A3C (Continuous)\cite{A3Ccont}    & 591 $\pm$ 45  \\
    World model with random MDN-RNN\cite{Corentin2018rnnfix}     &   870  $\pm$ 120 \\            
    World model (V model)\cite{DBLP:journals/corr/abs-1803-10122} &    632 $\pm$ 251\\
    World model\cite{DBLP:journals/corr/abs-1803-10122}     &   {\bf 906}  $\pm$ 21 \\
    GA\cite{DBLP:journals/corr/abs-1906-08857}     &   {\bf 903}  $\pm$  73 \\    
    \midrule 
    RCRC model (Visual model)  &    864  $\pm$  79\\     
    RCRC model     &  {\bf 901}  $\pm$  20 \\
    \bottomrule
  \end{tabular}
  \label{tab:table}
\end{table}

\subsection{Result}
The best scores among 16 workers are shown in Figure\ref{fig:fig4}. Each workers score are evaluated as average score over 8 randomly generated tracks. Although the world model improved score faster than the full RCRC model, the full RCRC model also reached high score. The full RCRC model reached an average score around 900 at 200 generations and stable high score after 400 generations, while the world model reached stable high score after 250 generations. This result shows that the full RCRC model is comparable to world model at the same condition, regardless of no training process in feature extractions. However, the full RCRC model was slower than the world model to achieve stable high scores.

Incredibly, the dense model reached an average score above 880 over 8 randomly generated tracks, and the visual model reached above 890. The dense model's score transition has higher volatility than the visual model's score transition. Furthermore, the visual model's score is less stable than the full RCRC model's score. These results shows that only one dense process can extract visual features even though the weight are random and fixed, and the features extracted by convolutional process and ESN improved scores. 

We also test the ability of single dense network in the MNIST datasets\cite{mnistdata} which is benchmark dataset of image recognition task, including 28$\times$28 gray-scaled handwritten images. The MNIST dataset contains 60000 training data and 10000 testing data. In experiments, we merged these data, and randomly sampled 60000 data as training data and 10000 data as testing data with no duplication to evaluate model ability by multiple datasets.  As input to the dense layer, we used 784 vector that is flatten representation of 28$\times$28 gray-scaled handwritten images, and set the dense layer has 512 units. Each input vector is divided by 255 to normalize value. The weight of the dense layer is randomly generated from a Gaussian distribution $\mathcal{N}(0, 0.06^2)$ and then fixed. After extracting features by the dense layer, we uses these features as input to logistic regression with L2 penalty to classify images into 10 classes. As a result of this experiments, we confirmed the feature extracted by random fixed-weight single dense layer has ability to achieve average accuracy score $91.58 \pm 0.27 (\%)$ over 20 trials by linear model.  Surprisingly, this result shows that only single dense layer with random fixed-weight has ability to extract visual features.

The generalization ability of the visual model and the full RCRC model that evaluated  as average score over 100 randomly generated tracks are shown in Table\ref{tab:table}. The full RCRC model achieved above 901$\pm$20 that is comparable to state of the art approaches such as the world model approach\cite{DBLP:journals/corr/abs-1803-10122} and GA approach\cite{DBLP:journals/corr/abs-1906-08857}. To achieve over 900 score of 100 randomly generated tracks, the models is only allowed to mistake a few driving. Therefore the full RCRC model can be regarded as having ability to solve {\tt CarRacing-v0}. 

Although the visual model extracts 512-dimensional visual features with random fixed-weight CNN, it achieves 864$\pm$79 which is better than the V model that uses only 32-dimensional features extracted by VAE as input to controller layer in world model. Furthermore, the time-series features extracted by ESN improves driving. 


\section{Discussion and Future work}

In this study, we focused on extracting features that sufficiently express the environment state, rather than those that are trained to solve the RL task. To this end, we developed a novel model called RCRC, which, using random fixed-weight CNN and a novel ESN method, respectively, extracts visual features from environment state pixels and time-series features from a series of the state pixels. Therefore, no training process is required, and features can be efficiently extracted. In the controller layer, a linear combination of both features and actions is trained using CMA-ES. This model architecture results in highly practical features that omit the training process and reduce computational costs, and there is no need to store large volumes of data. We also show that RCRC achieves state of the art scores in a popular continuous RL task, {\tt CarRacing-v0}. This result brings us to the conclusion that network structures themselves, such as CNN and ESN, have the capacity to extract features. 

We also found that the single dense network and simple CNN model with random fixed-weight can extract visual features, and these models achieved high scores. Although VAE has desirable features such as ability to reconstruct the input and high interpretability of latent space by using reparameterization trick which uses Gaussian noise to use backpropagation, we consider that large definitive features can also extract expressive enough visual features. 

Because of our limited computing resources, we were unable to assign more workers to CMA-ES. There is a possibility that more efficient and stable training could be performed by assigning more workers. Although RCRC can take wide range of actions and parallelization by using CMA-ES, it is not suitable for the task that is hard to real simulation because it has to evaluate parameters by real simulation. 

As a further improvement, there is a possibility that the score can be improved and made it more stable by using a multi-convolutional reservoir computing layer to extract multiple features\cite{massar2013mean}. The current convolutional reservoir computing layer uses random weight samples generated from Gaussian distributions. Therefore, it can easily obtain multiple independent features by using different random seeds.

Our results have the potential to make RL widely available. Recently, many RL models have achieved high accuracy in various tasks, but most of them have high computational costs and often require significant time for training. This makes the introduction of RL inaccessible for many. However, by using our RCRC model, anyone can train the model at a high speed with much lower computational costs, and importantly, anyone can build a highly accurate model. In addition, RCRC can handle both continuous- and discrete-valued tasks, and even when the environment changes, training can be performed without any prior learning such as the VAE and MDN-RNN in world model. Therefore, it can be used easily by anyone in many environments.

In future work, we consider making predictions from previous extracted features and actions to the next ones to be an important and promising task. Because the ESN was initially proposed to predict complex time-series, it can be assumed to have capacity to predict future features. If this prediction is achieved with high accuracy, it can self-simulate RL tasks by making iterative predictions from initial state pixels. This will help to broaden the scope of RL applications.

\section{Acknowledgements}

The authors are grateful to Takuya Yaguchi for the discussions on reinforcement learning. We also thank Hiroyasu Ando for helping us to improve the manuscript.

\bibliographystyle{unsrt}

\end{document}